\documentclass[10pt,twocolumn,letterpaper]{article}
\usepackage{wacv}
\usepackage{amsmath,amssymb,bm}
\usepackage{booktabs,multirow}
\usepackage{graphicx}
\usepackage{xcolor,microtype}

\newcommand{\ours}{PsyDrift}
\newcommand{\ddpm}{DDPM}
\newcommand{\vit}{ViT}

\newcommand{\loso}{LOSO}
\newcommand{\auroc}{AUROC}
\newcommand{\asd}{ASD}
\newcommand{\hc}{HC}
\definecolor{wacvblue}{rgb}{0.21,0.49,0.74}
\usepackage[pagebackref,breaklinks,colorlinks,allcolors=wacvblue]{hyperref}
\hypersetup{
  pdfauthor={Yasaman Mohammadpour, Pouya Shaeri, Ariane Middel},
  pdftitle={PsyDrift: Healthy-Reference Latent Diffusion for Anomaly Scoring in Autism Spectrum Disorder}
}
\def\wacvPaperID{9}
\def\confName{WACV}
\def\confYear{2027}
\title{\vspace{-0.5cm}PsyDrift: Healthy-Reference Latent Diffusion for
Anomaly Scoring in\\Autism Spectrum Disorder}

\author{
Yasaman Mohammadpour\\
\textit{School of Computing and}\\
\textit{Augmented Intelligence}\\
\textit{Arizona State University}\\
\textit{Tempe, Arizona, USA}\\
{\tt\small ymoham15@asu.edu}
\and
Pouya Shaeri\\
\textit{School of Computing and}\\
\textit{Augmented Intelligence}\\
\textit{Arizona State University}\\
\textit{Tempe, Arizona, USA}\\
{\tt\small pshaeri@asu.edu}
\and
Ariane Middel\\
\textit{The GAME School}\\
\textit{Arizona State University}\\
\textit{Tempe, Arizona, USA}\\
{\tt\small ariane.middel@asu.edu}
}

\begin{document}
\maketitle
\begingroup
\renewcommand{\thefootnote}{}
\footnotetext{%
Extended version of a paper accepted at the CVPath-Vision4Neuro
Workshop at WACV 2027.
\copyright~2027 IEEE. Personal use of this material is permitted.
Permission from IEEE must be obtained for all other uses, in any
current or future media, including reprinting/republishing this
material for advertising or promotional purposes, creating new
collective works, for resale or redistribution to servers or lists,
or reuse of any copyrighted component of this work in other works.%
}
\endgroup
\begin{abstract}
Individual heterogeneity and acquisition-site variability complicate neuroimaging-based characterization of autism spectrum disorder (ASD). We propose \ours{}, combining ASD/healthy-control-supervised functional-connectivity representations with healthy-reference latent diffusion. An encoder produces 128-dimensional representations, and a denoising diffusion probabilistic model (\ddpm{}) learns exclusively from healthy training latents. Distance from a diffusion-denoised reconstruction provides a continuous deviation score. We evaluate 1,102 subjects from all 20 ABIDE-I sites using five-seed leave-one-site-out cross-validation. The classifier and \ddpm{} achieve mean-site \auroc{}s of $0.798 \pm 0.003$ and $0.718 \pm 0.013$, respectively; uncertainties denote seed standard deviations. Matched healthy-reference scorers on the same representation achieve 0.68--0.71; the statistical significance of diffusion's modest numerical advantage remains unestablished. Within ASD, scores show weak associations with ADOS Total (Spearman $\rho = 0.108$, $p = 0.038$) and full-scale IQ ($\rho = -0.138$, $p = 0.002$), without adjustment for age, sex, site, or head motion. Frozen evaluation without retraining or hyperparameter adaptation yields pooled AUROCs of 0.69 on ABIDE-II and 0.61 on ADHD-200, providing preliminary evidence of cross-cohort and cross-disorder transfer. These results demonstrate healthy-reference deviation scoring within an ASD-supervised representation; incremental predictive and clinical value remains unestablished.

\noindent Code: \href{https://github.com/yasamanmohammdpour/PsyDrift}{github.com/yasamanmohammdpour/PsyDrift}
\end{abstract}
\section{Introduction}
\label{sec:intro}

\begin{figure*}[t]
  \centering
  \includegraphics[width=\linewidth]{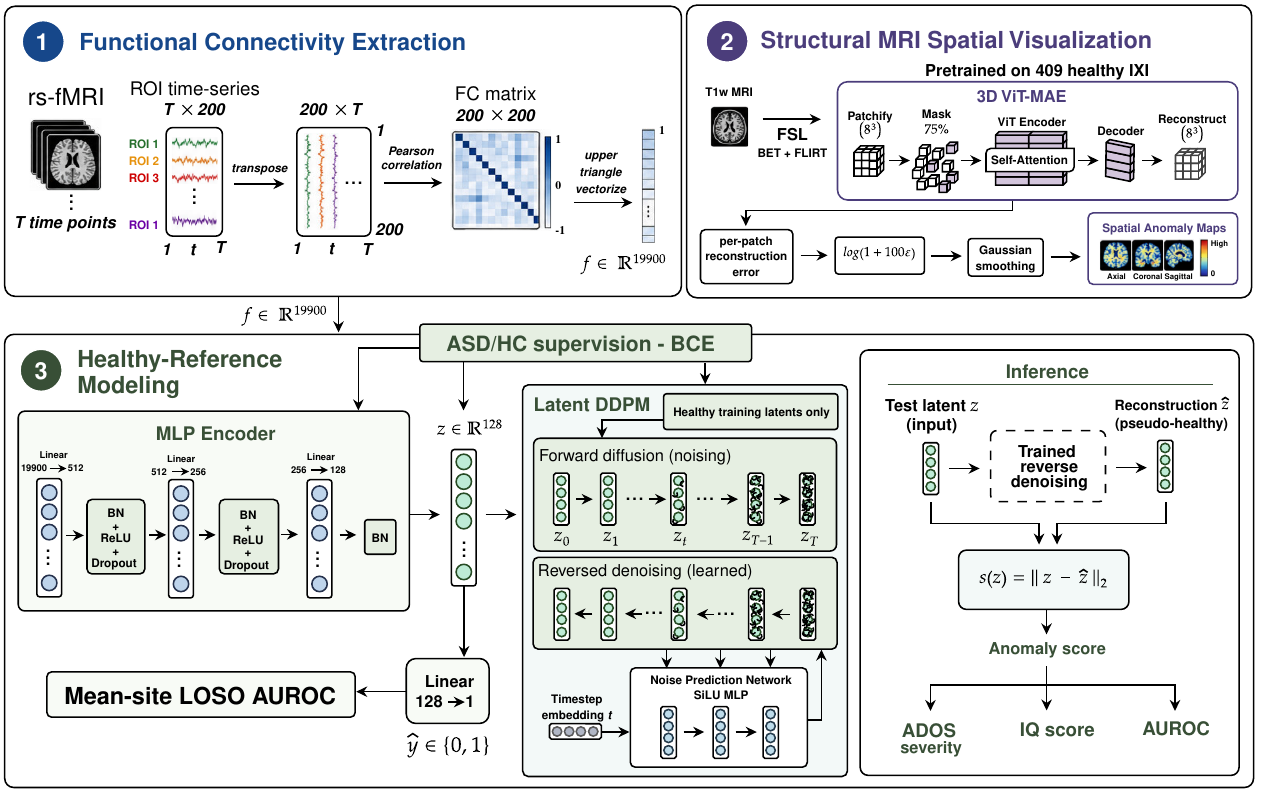}
  \caption{\textbf{\ours{} framework.}
(\textbf{1}) \textit{FC extraction}: ROI time-series are transposed
from $T{\times}\mathrm{ROI}$ to $\mathrm{ROI}{\times}T$ and
Pearson-correlated into a $200{\times}200$ FC matrix, whose upper
triangle, excluding the diagonal, yields
$\mathbf{f}\in\mathbb{R}^{19900}$.
(\textbf{2}) \textit{Exploratory structural visualization}:
FSL-preprocessed T1w volumes pass through a 3D \vit{} pretrained
via MAE on 409 healthy IXI subjects. Log-scaled reconstruction
errors yield structural maps independent of the fMRI classifier
and DDPM score.
(\textbf{3}) \textit{Healthy-reference modeling}: a diagnostically
supervised MLP maps FC features to $\mathbf{z}\in\mathbb{R}^{128}$.
A latent \ddpm{} trained only on healthy training latents produces
$\hat{\mathbf{z}}$, with anomaly score
$s(\mathbf{z})=\|\mathbf{z}-\hat{\mathbf{z}}\|_2$
quantifying deviation from the learned healthy reference.}
  \label{fig:architecture}
\end{figure*}

Mental disorders are a leading cause of disability worldwide~\cite{who2022},
accounting for approximately one-third of years lived with disability
in a global burden re-estimation~\cite{vigo2016estimating}.
The CDC's 2023 surveillance report identified autism spectrum disorder
(\asd{}) in approximately 1 in 36 eight-year-old children in monitored
US communities~\cite{cdcasd2023}.
ASD involves persistent social-communication challenges, restricted
repetitive behaviors, and highly variable symptom severity.
Diagnosis relies on behavioral assessments such as the Autism Diagnostic
Observation Schedule (ADOS), which require specialized expertise and time
and are subject to inter-rater variability.

Resting-state functional MRI (rs-fMRI) offers a non-invasive means of
characterizing \asd{}, revealing altered functional connectivity in
networks governing social cognition, language, and executive
function~\cite{di2014autism}.
Models trained on functional connectivity (FC) matrices show meaningful
classification performance, but diagnostic-label dependence and
acquisition-site variability complicate their translation.
Supervised classifiers require diagnostic labels that can be costly to
obtain, inconsistent across sites, or unavailable for some records.
Here, we investigate healthy-reference scoring within a diagnostically
supervised representation to quantify deviation from controls alongside
diagnostic prediction. This approach retains label dependence in
representation learning. Semi-supervised learning provides a complementary direction for reducing
dependence on diagnostic annotations. In another application domain,
Shaeri and Katanforoush~\cite{shaeri2023semi} combine pretrained sentiment
representations with a self-attentive LSTM and semi-supervised
self-training for fake-news detection. This motivates investigating
whether pretrained representations and carefully selected pseudo-labels
could support future connectome encoders when diagnostic annotations
are limited.

Multi-site fMRI studies such as ABIDE-I~\cite{di2014autism} span 20
international sites with heterogeneous scanners, acquisition protocols,
and demographics, challenging generalization to held-out
sites~\cite{fortin2018harmonization}.
Leave-one-site-out (\loso{}) cross-validation directly evaluates
performance on unseen acquisition sites. Random subject splits share
sites across training and testing and therefore address a different,
potentially less challenging generalization setting.

We propose \ours{}, a healthy-reference latent diffusion framework
for anomaly scoring in \asd{} (Figure~\ref{fig:architecture}).
A supervised MLP maps FC to a diagnostically structured representation;
a latent DDPM~\cite{ho2020denoising} fitted only to healthy training
latents provides a continuous reconstruction-based deviation score.
The encoder uses ASD/HC labels, so the complete pipeline remains
diagnostically supervised. We compare diffusion with simpler
healthy-reference scorers on the same representation to distinguish
scoring effects from representation effects. Our contributions are fourfold. (i) We combine supervised FC
representations with healthy-reference latent diffusion.
(ii) We compare the diffusion scorer with six matched healthy-only
alternatives, assess corruption-step sensitivity, and ablate supervised
versus reconstruction-based representation learning.
(iii) We evaluate all 20 ABIDE-I sites across five seeds and test
frozen-model transfer to ABIDE-II and, exploratorily, ADHD-200.
(iv) We examine exploratory clinical associations after adjusting for
classifier logit and document an FC-orientation error in our pipeline,
quantifying its effect on performance.
\section{Related Work}
\label{sec:related}

\subsection{Functional Connectivity-based ASD Detection}

Early ASD classifiers applied support vector machines and elastic-net
regularization to vectorized rs-fMRI connectivity~\cite{di2014autism}.
Studies document altered default-mode, social, and attention
networks~\cite{padmanabhan2017default,farrant2016atypical,ilioska2023connectome},
supporting rs-fMRI-based psychiatric biomarker research~\cite{canario2021review,van2010exploring}.

BrainNetCNN~\cite{kawahara2017brainnetcnn} uses graph-structured
convolutions. BNT~\cite{kan2022brain} combines ROI-level self-attention
with Orthonormal Clustering Readout, while
Com-BrainTF~\cite{bannadabhavi2023community} models intra- and
inter-community interactions through hierarchical local-global attention.
Rahman~\etal~\cite{rahman2026multimodal} fuse functional and structural
MRI in a population graph using asymmetric cross-attention, reporting
82\% accuracy under \loso{} on 17 quality-filtered ABIDE-I sites.
Other graph-based approaches include BrainGNN~\cite{li2021braingnn},
BrainGB~\cite{cui2022interpretable}, and FBNetGen~\cite{kan2022fbnetgen}.

These methods primarily optimize diagnostic prediction. We additionally
study healthy-reference deviation and compare it with continuous
classifier logits under matched all-site LOSO evaluation.

\subsection{Normative Modeling in Autism}

Normative modeling of ASD from typically developing reference populations
has been explored for structural and functional features. Shan~\etal~\cite{shan2022mapping}
model individualized structural deviations and identify heterogeneous ASD
phenotypes. Liu~\etal~\cite{liu2025identifying} build normative models from
static and dynamic functional connectivity, connecting functional subtypes
with behavioral profiles. Building on these healthy-reference approaches, \ours{} studies latent
diffusion on ASD-supervised representations against matched simpler scorers.

\subsection{Anomaly Detection with Diffusion Models}

Denoising diffusion probabilistic models~\cite{ho2020denoising} have
become the state of the art for generative modeling and have
recently been applied to medical image anomaly detection.
AnoDDPM~\cite{wyatt2022anoddpm} uses simplex noise and partial
diffusion to detect brain tumors and lesions through
pixel-space reconstruction error.
Latent diffusion models~\cite{rombach2022high} perform generation
in compressed representation spaces, enabling faster inference
and richer latent structure.
We investigate latent \ddpm{} scoring of functional-connectivity
embeddings for ASD.
We distinguish representation and scorer effects through separate
ablations (Sections~\ref{sec:matched} and~\ref{sec:ablation}).

\subsection{Vision Transformers for Brain MRI}

Masked autoencoders (MAE)~\cite{he2022masked} have demonstrated strong
representation learning from brain volumes, learning anatomically
meaningful features from large-scale pretraining.
BrainIAC~\cite{tak2026generalizable} is a ViT-B/16 pretrained with SimCLR
on structural brain MRI scans, published in Nature Neuroscience 2026.
Our IXI-trained 3D ViT provides independent exploratory structural
visualizations; a paired-subset ablation separately assesses structural features. Human feedback provides a complementary mechanism for refining
image-based models. Shaeri~\etal~\cite{shaeri2025explainable} investigate
interactive segmentation in which targeted human corrections provide
critic feedback signals. This feedback structure motivates future
expert-guided review of structural reconstructions and preprocessing
failures in \ours{}, with corrections evaluated for their effects on
representation quality and cross-site performance.

\subsection{Cross-site Generalization}

Scanner-induced variation confounds multi-site fMRI
classification~\cite{fortin2018harmonization}.
Random splits allow site-specific patterns to inflate reported
performance~\cite{kan2022brain,bannadabhavi2023community}; \loso{} is the rigorous
alternative for cross-site testing. Acquisition heterogeneity also motivates evaluation across datasets and
devices in wearable sensing. HeteroWear~\cite{mohammadpour2026heterowear}
combines sensor-aware self-supervised representation learning with
leave-one-dataset-out pretraining and held-out-device evaluation.
Its separation of within-sensor temporal modeling from cross-sensor
interactions offers an architectural direction for future extensions
of \ours{} that combine connectivity, structural, and physiological
measurements. Our primary ABIDE-I evaluation uses
\loso{}, complemented by frozen-model external-cohort tests.

\section{\ours{}: Method}
\label{sec:method}

\begin{figure*}[t]
  \centering
\includegraphics[width=\linewidth,trim=0 0 0 25,clip]{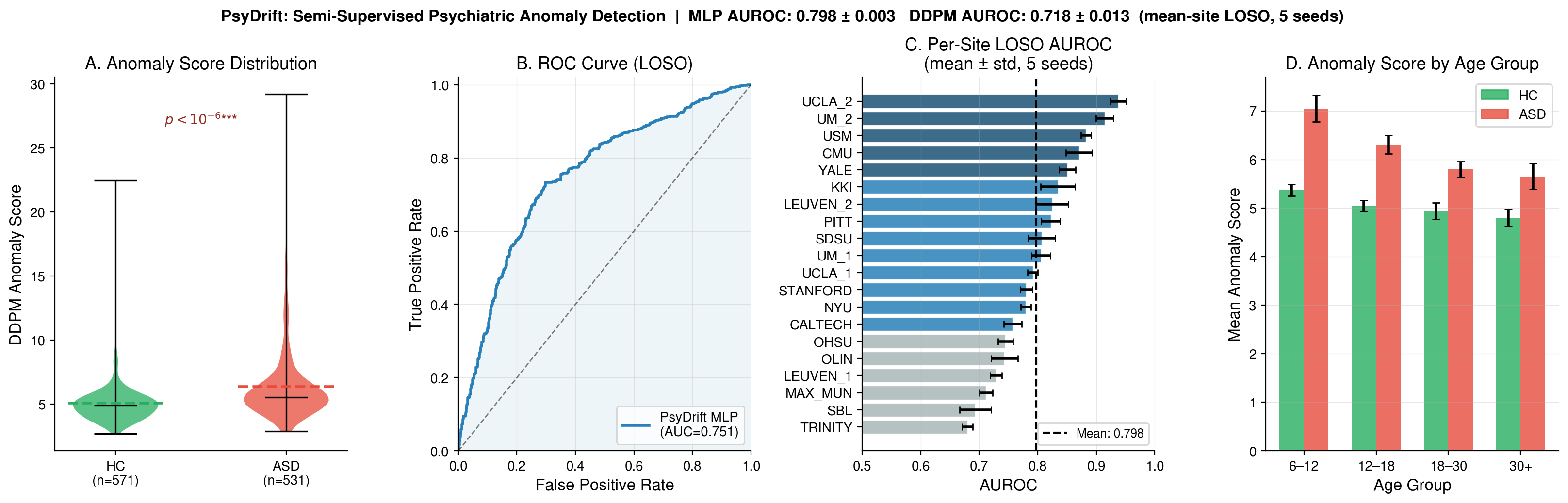}
  \caption{\textbf{Main Results.}
  (\textbf{A}) \ddpm{} anomaly score violin plots: \asd{} subjects show
  significantly higher scores than \hc{} ($p{<}10^{-6}$, Mann-Whitney $U$).
  (\textbf{B}) MLP ROC curve from pooled out-of-fold predictions
across ABIDE-I sites for seed 42 (pooled \auroc{} = 0.751;
mean-site \auroc{} = $0.798\pm0.003$ across five seeds).
  (\textbf{C}) Per-site \auroc{} with mean $\pm$ std across 5 seeds;
  12 of 20 sites exceed 0.75.
  (\textbf{D}) \ddpm{} anomaly scores by age group; \asd{} consistently
  exceeds \hc{} across all groups.}
  \label{fig:main_results}
\end{figure*}

\begin{figure}[t]
  \centering
  \includegraphics[width=\linewidth]{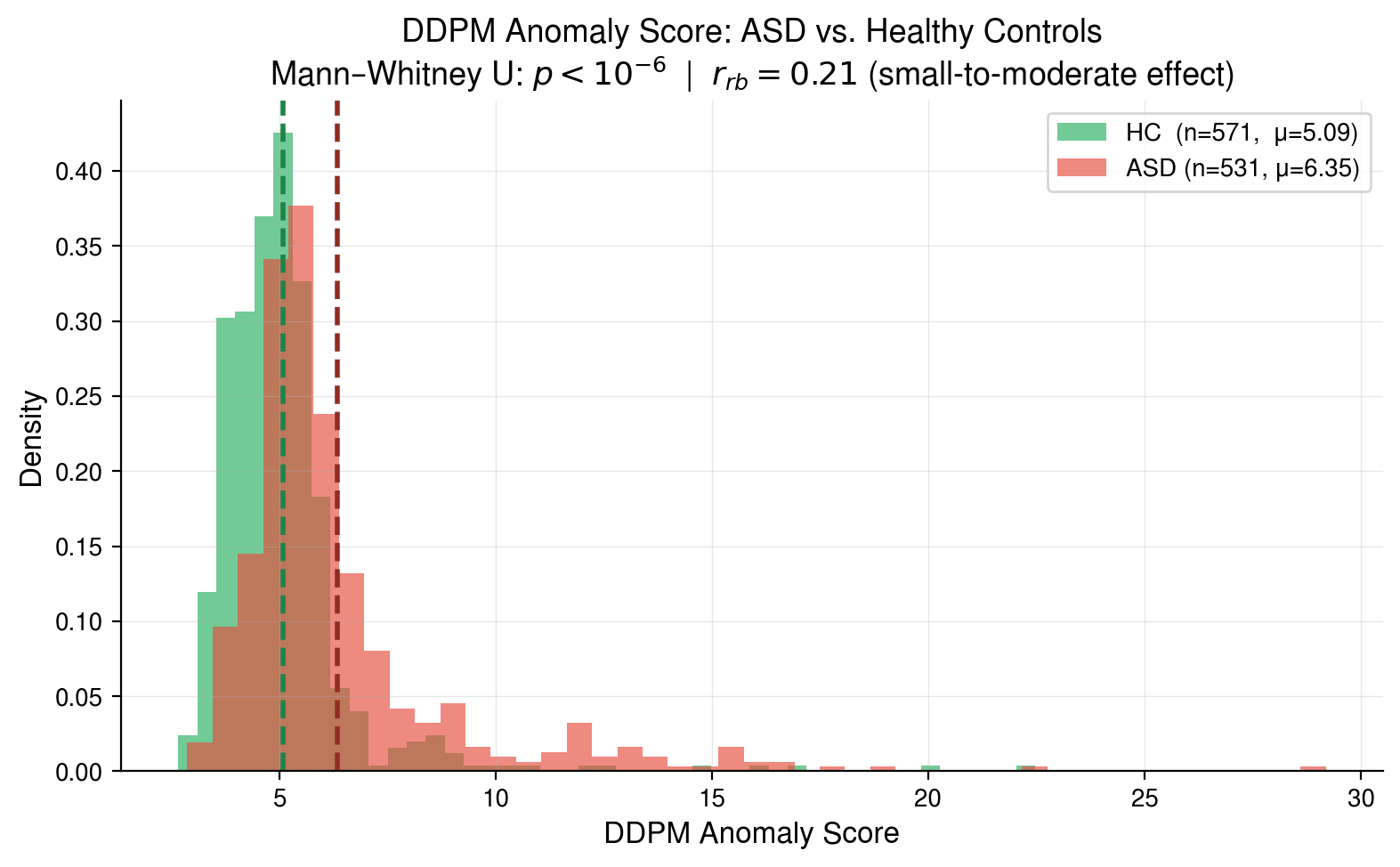}
  \caption{\textbf{\ddpm{} Anomaly Score Distribution (Seed 42).}
  \asd{} (red, $\mu{=}6.35$) shows a substantially heavier right tail
  than \hc{} (green, $\mu{=}5.09$), indicating a subgroup of \asd{}
  subjects with extreme deviation from the healthy manifold.
  Across-subject group summaries and the statistical test are reported
  separately in Table~\ref{tab:anomaly}.}
  \label{fig:anomaly_dist}
\end{figure}

\begin{figure*}[t]
  \centering
  \includegraphics[width=\linewidth]{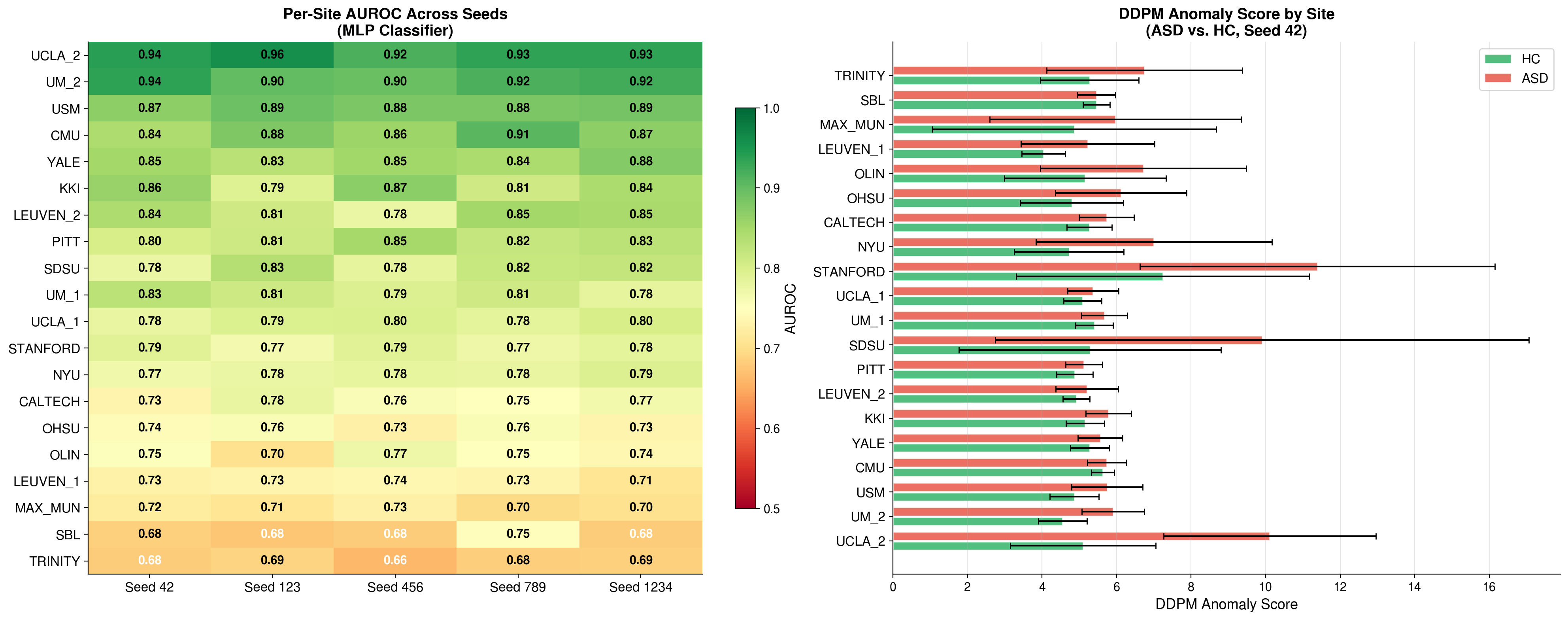}
  \caption{\textbf{Cross-Site Analysis.}
  (\textbf{Left}) Site-level MLP AUROC across five seeds.
  UCLA\_2 (0.94) and UM\_2 (0.93) are top performers;
  SBL (0.68) and TRINITY (0.68) are most challenging.
  (\textbf{Right}) \ddpm{} anomaly scores by site for \asd{} vs.\ \hc{};
  \asd{} scores exceed \hc{} at the majority of sites.}
  \label{fig:site_analysis}
\end{figure*}

\begin{figure}[t]
  \centering
  \includegraphics[width=\linewidth]{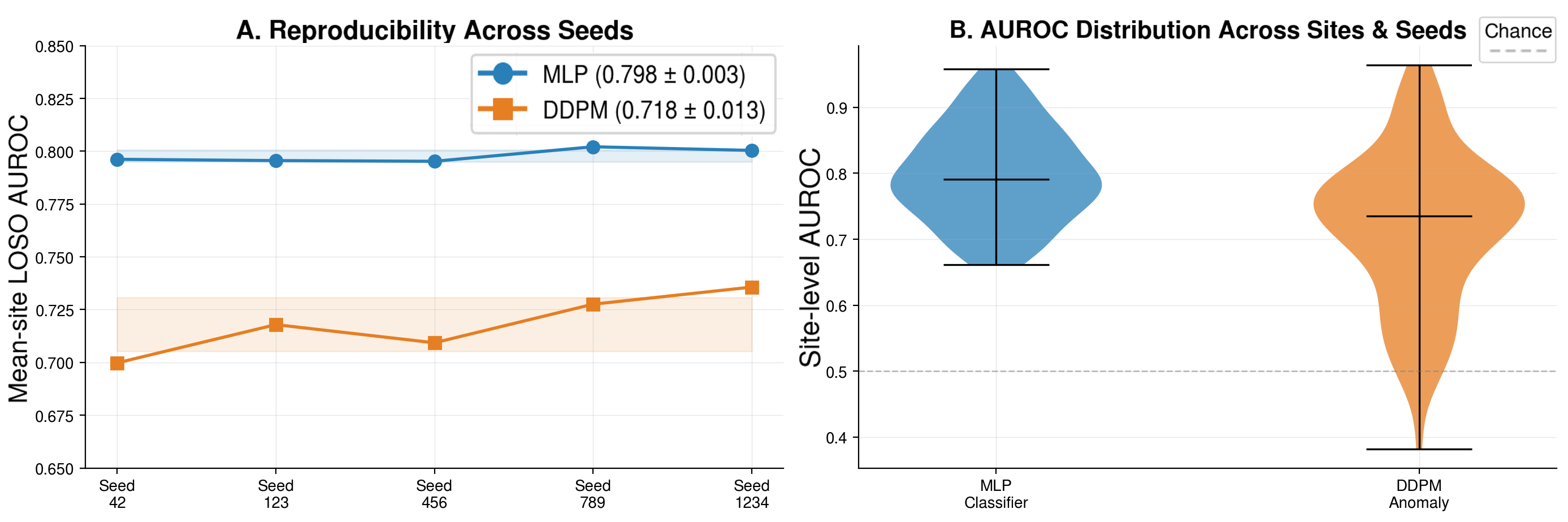}
  \caption{\textbf{Seed-Level Reproducibility.}
  (\textbf{A}) Mean-site \loso{} \auroc{} across 5 seeds for the MLP
  classifier ($0.798\pm0.003$, blue) and \ddpm{} anomaly detector
  ($0.718\pm0.013$, orange).
  (\textbf{B}) Site-level \auroc{} distributions across all 20 sites
  and 5 seeds; the MLP distribution is concentrated above 0.7 while
  the \ddpm{} has a wider spread reflecting site heterogeneity.}
  \label{fig:seed_stability}
\end{figure}

\begin{figure}[t]
  \centering
  \includegraphics[width=\linewidth]{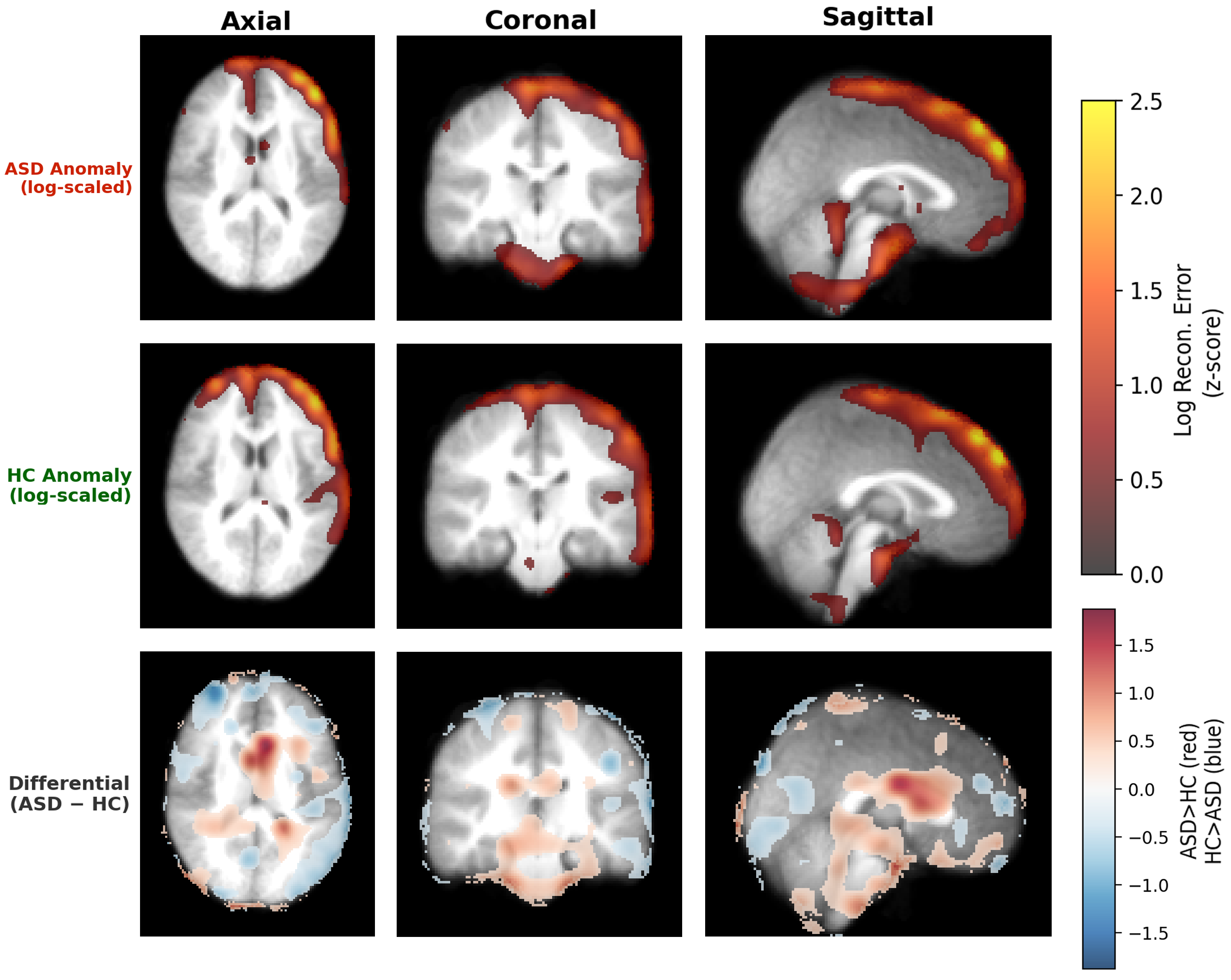}
  \caption{\textbf{Independent Exploratory Structural Maps (Group Level).}
  Log-scaled 3D \vit{} reconstruction error for \asd{} (top) and
  \hc{} (middle) group means, and differential map ASD$-$HC (bottom).
  Red regions in the differential map indicate consistently higher
  reconstruction error in \asd{}, concentrated in frontal and
  temporal-parietal cortices. These structural maps do not contribute to
  or spatially explain the fMRI DDPM score.
  }
  \label{fig:brain_maps}
\end{figure}

\begin{figure}[t]
  \centering
  \includegraphics[width=\linewidth]{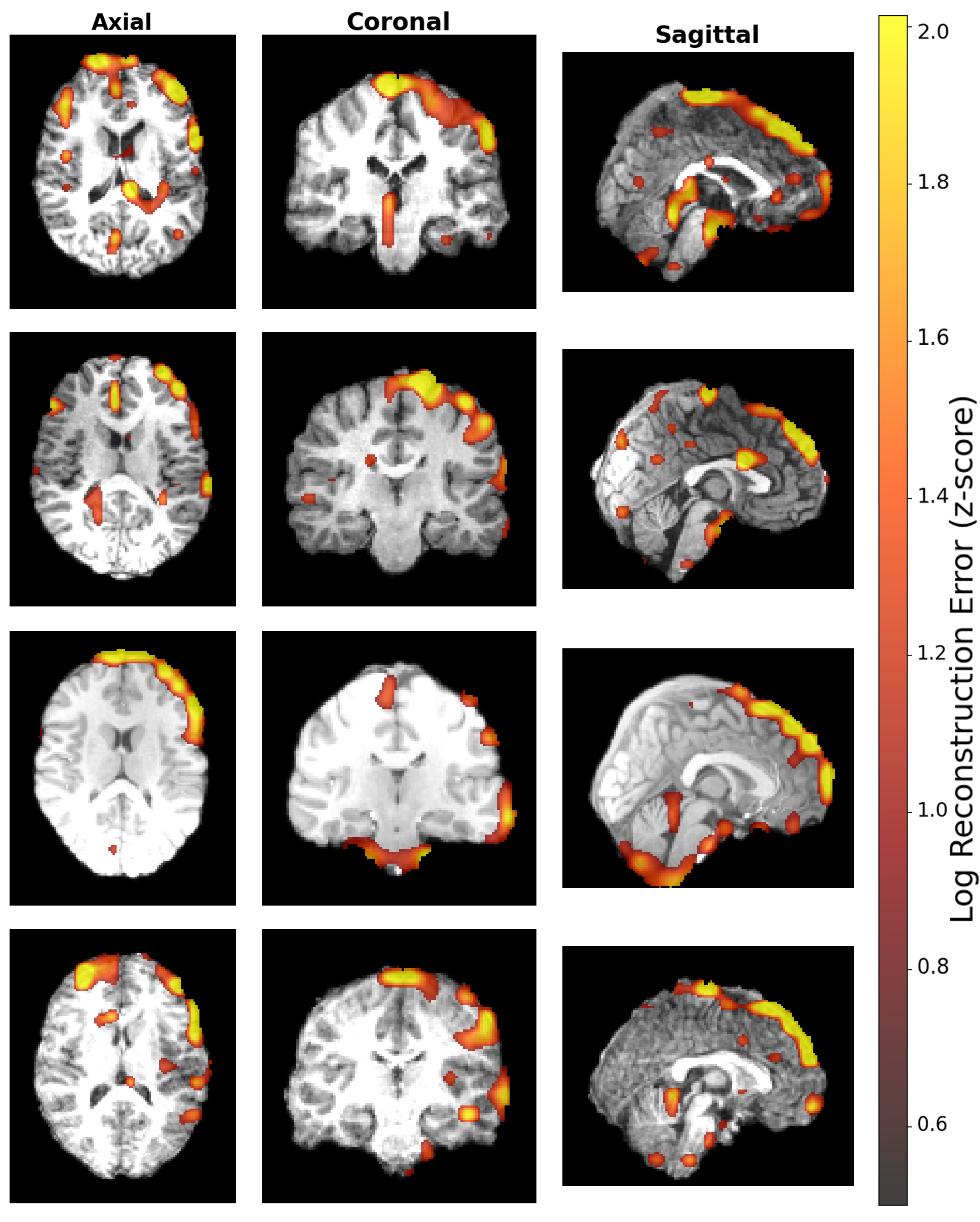}
  \caption{\textbf{Independent Structural Maps for Selected ASD Cases.}
  Four \asd{} subjects with the highest \ddpm{} anomaly scores.
  Despite phenotypic heterogeneity, consistent anomaly patterns
  emerge in frontotemporal and parietal regions.
  \ddpm{} anomaly scores are shown per subject (top to bottom:
  highest to lowest among top-4). The maps are generated by the separate
  structural model and are not explanations of the fMRI score.}
  \label{fig:individual_maps}
\end{figure}

Figure~\ref{fig:architecture} summarizes FC extraction, supervised MLP
encoding, healthy-reference DDPM scoring, and an independent structural
visualization branch. Supervision structures the latent space; the DDPM
models its healthy-control subset.

\subsection{Functional Connectivity Features}
\label{sec:fc}

Each subject's rs-fMRI data is parcellated using the CC200
atlas~\cite{craddock2012whole} into $N{=}200$ region-of-interest (ROI)
time-series.
Given the time-series matrix in its native file format
($T_i \times N$), we always transpose to obtain
$\mathbf{T} \in \mathbb{R}^{N \times T_i}$ before computing
Pearson correlation.
Section~\ref{sec:preprocessing} quantifies the impact of incorrect orientation.
After z-scoring each ROI independently, the Pearson correlation
matrix $\mathbf{C} \in \mathbb{R}^{N \times N}$ is computed,
and its upper triangle (excluding the diagonal) yields the feature
vector $\mathbf{f} \in \mathbb{R}^{19900}$.

\subsection{Supervised MLP Encoder}
\label{sec:encoder}

We project $\mathbf{f}$ to a latent representation
$\mathbf{z} \in \mathbb{R}^{128}$ via a three-layer MLP
with batch normalization and dropout ($p{=}0.5$):
\begin{align}
  \mathbf{h}_1 &= \text{BN}(\text{ReLU}(\mathbf{W}_1\mathbf{f}+\mathbf{b}_1)),\nonumber\\
  \mathbf{h}_2 &= \text{BN}(\text{ReLU}(\mathbf{W}_2\mathbf{h}_1+\mathbf{b}_2)),\nonumber\\
  \mathbf{z}   &= \text{BN}(\mathbf{W}_3\mathbf{h}_2+\mathbf{b}_3).
\end{align}
A linear head $\hat{y} = \mathbf{w}^\top\mathbf{z}$ produces
the binary \asd{} logit.
The encoder and head are trained end-to-end with binary
cross-entropy under \loso{}, organizing the latent space around
functionally discriminative features.
Section~\ref{sec:ablation} compares this representation with an
unsupervised reconstruction-trained encoder.

\subsection{Latent DDPM for Anomaly Scoring}
\label{sec:ddpm}

After training the encoder, we fit a \ddpm{} on the set
$\mathcal{Z}_\text{HC} = \{\mathbf{z}_i : y_i = \text{HC}\}$
of healthy latent vectors from the training fold only.
We freeze the ASD/HC-supervised encoder before fitting the DDPM
and the matched scorers on training-fold control latents.

\noindent\textbf{Forward process.}
A Markov chain gradually corrupts $\mathbf{z}_0 \in \mathcal{Z}_\text{HC}$
with Gaussian noise:
\begin{equation}
  q(\mathbf{z}_t|\mathbf{z}_0)
  = \mathcal{N}\!\left(
    \sqrt{\bar\alpha_t}\,\mathbf{z}_0,\,
    (1{-}\bar\alpha_t)\mathbf{I}
  \right),
\end{equation}
where $\bar\alpha_t{=}\prod_{s=1}^{t}(1{-}\beta_s)$ accumulates
the $T{=}300$ linear noise schedule
($\beta_1{=}10^{-4}$, $\beta_T{=}0.02$).

\noindent\textbf{Reverse process.}
A three-layer MLP with SiLU activations and timestep embedding
$\mathbf{e}_t\!\in\!\mathbb{R}^{32}$ learns to predict added noise:
\begin{equation}
  \mathcal{L}_{\ddpm{}}
  = \mathbb{E}_{t,\mathbf{z}_0,\boldsymbol\epsilon}\!\left[
    \bigl\|\boldsymbol\epsilon
    - \hat{\boldsymbol\epsilon}_\theta(\mathbf{z}_t, t)\bigr\|^2
  \right],
  \quad \mathbf{z}_0 \in \mathcal{Z}_\text{HC}.
\end{equation}

\noindent\textbf{Anomaly score.}
At inference, a test subject's latent $\mathbf{z}$ is partially
diffused to step $t^*{=}100$ and denoised back to step 0,
yielding a latent reconstruction $\hat{\mathbf{z}}$.
The anomaly score is:
\begin{equation}
  s(\mathbf{z}) = \|\mathbf{z} - \hat{\mathbf{z}}\|_2.
  \label{eq:anomaly}
\end{equation}
Higher scores indicate larger reconstruction errors relative
to the learned healthy reference; they are not calibrated
measures of disease severity.
This score is computed identically for \asd{} and \hc{} test subjects
using only the \ddpm{} trained on healthy training subjects.

\subsection{Exploratory Structural Visualization}
\label{sec:vit}

To provide an independent exploratory view of structural
reconstruction differences alongside the scalar fMRI anomaly score,
we pretrain a 3D \vit{} via MAE~\cite{he2022masked} on 409 healthy
IXI brain volumes only, with no ABIDE-I subjects used for pretraining.
Volumes are skull-stripped and registered to MNI152 1mm space
using FSL~\cite{jenkinson2012fsl} and resized to $112{\times}136{\times}112$.
The \vit{} uses patch size $8^3$, embedding dimension 384,
6 Transformer layers, 6 attention heads, and masking ratio 75\%,
achieving MAE loss of 0.0065 at convergence.

\noindent\textbf{Spatial anomaly maps.}
The per-patch reconstruction error $\epsilon_p$ is mapped to 3D
and log-transformed:
\begin{equation}
  \tilde\epsilon_p = \log(1 + 100\cdot\epsilon_p).
\end{equation}
The log transform reveals spatial variation obscured by the low dynamic
range of raw reconstruction errors.
We apply Gaussian smoothing ($\sigma{=}2.5$ voxels) and z-score
normalization within the brain mask.
These exploratory structural maps are computed separately from
the primary FC classifier and DDPM score and are not spatial
attributions of that score.

\subsection{Training Protocol}
\label{sec:training}

For each \loso{} fold: (1) train the MLP encoder with AdamW
($\eta{=}10^{-3}$, $\lambda{=}10^{-2}$), cosine annealing,
up to 150 epochs, and early stopping (patience 25);
(2) fit the \ddpm{} on training-fold healthy latents with Adam
($\eta{=}10^{-3}$) for 50 epochs;
(3) compute classification probabilities from the supervised logit
and anomaly scores from Eq.~\eqref{eq:anomaly} on test subjects.
Experiments run on a single NVIDIA A100 80GB GPU using PyTorch 2.5.
The primary encoder/DDPM pipeline uses five random seeds
(42, 123, 456, 789, 1234); reported $\pm$ values denote seed standard
deviations where available.
These standard deviations summarize variability across complete runs
and do not separately quantify inference-time stochastic variability.
\section{Experimental Setup}
\label{sec:experiments}

\subsection{Dataset}

We evaluate on ABIDE-I~\cite{di2014autism}, a publicly available
multi-site rs-fMRI dataset for \asd{} research.
After phenotypic filtering, we retain 1,102 subjects
(531 \asd{}, 571 \hc{}) from 20 international sites
spanning ages 6--64 years.
FC features use the CC200 extraction in Section~\ref{sec:fc}.
Of these 1,102 subjects, 980 have accompanying T1-weighted
structural MRI, which we preprocess using FSL BET (skull stripping,
$f{=}0.3$) and FLIRT (12-DOF affine registration to MNI152 1mm
space) for the ViT visualization component.
The 409 preprocessed IXI volumes are reserved for ViT pretraining.

\subsection{Evaluation Protocol}
\label{sec:protocol}

We adopt leave-one-site-out (\loso{}) cross-validation:
for each of the 20 ABIDE-I sites, the model is trained on subjects
from all 19 remaining sites and evaluated on subjects from the
held-out site.
This separates training and evaluation sites; random subject splits
address a different generalization setting because acquisition sites
can be shared across partitions.
For each seed, mean-site \auroc{} is the unweighted mean of
the 20 held-out-site AUROCs.
We report the mean and standard deviation across five seeds
where available, alongside available classification accuracies.
External evaluation uses pooled \auroc{}, computed across subjects
within each external cohort.
Mean-site and pooled AUROCs use different aggregation procedures
and are not directly interchangeable.

\subsection{Baselines}

\noindent\textbf{LR + FC.}
PCA (200 components) applied to corrected Pearson FC features,
followed by $\ell_2$-regularized logistic regression ($C{=}0.01$).

\noindent\textbf{SVM + FC.}
RBF kernel SVM ($C{=}1.0$) on PCA-reduced, standardized FC features.

\noindent\textbf{BNT}~\cite{kan2022brain}.
Brain Network Transformer retrained on our ABIDE-I data using
the authors' public code, evaluated under \loso{}.

\noindent\textbf{Com-BrainTF}~\cite{bannadabhavi2023community}.
Community-aware hierarchical Transformer, trained with the
authors' public code under a random 70/10/20 split.
This different-protocol result is reported separately from
the \loso{} comparisons.

\noindent\textbf{Unsupervised AE + \ddpm{}.}
An FC autoencoder is trained with reconstruction loss only,
without diagnostic supervision, followed by our \ddpm{}.
This ablation compares reconstruction-based and diagnostically
supervised representation learning.

\noindent\textbf{Matched healthy-only scorers.}
To compare healthy-reference scoring methods, we hold the
128-dimensional supervised representation fixed and fit six
alternatives on training-fold controls only:
Euclidean distance to the healthy centroid;
Mahalanobis distance under a single Gaussian;
$k$NN distance with $k=10$;
one-class SVM;
VAE latent reconstruction error;
and normalizing-flow likelihood-based scoring.
These scorers use the same held-out subjects and mean-site
\loso{} metric as the DDPM.

\noindent\textbf{Clinical comparisons and sensitivity.}
Within ASD, we compare Spearman correlations of classifier logits
and DDPM scores with ADOS Total ($n=375$) and full-scale IQ
($n=498$), reporting 95\% confidence intervals.
Partial correlations assess DDPM-score associations after
controlling for classifier logit.
These analyses are exploratory and do not adjust for age, sex,
head motion, or acquisition site; they do not establish
incremental predictive or clinical utility.
We also evaluate corruption levels
$t^{\ast}\in\{25,50,100,150,200\}$ under mean-site \loso{}.
\section{Results}
\label{sec:results}

ABIDE-I AUROCs average held-out sites; means $\pm$ SD across five
seeds are reported where available. External AUROCs pool subjects
within each cohort and are not directly comparable to mean-site AUROC.

\subsection{ASD Classification}
\label{sec:cls_results}

Under all-20-site \loso{} (Table~\ref{tab:classification}), the MLP
achieves $0.798\pm0.003$ AUROC, numerically 0.097 above retrained
BNT; DDPM achieves $0.718\pm0.013$.
The MLP varies little across five runs, which does not quantify
site-level uncertainty or establish statistical superiority.
Table~\ref{tab:context} separately reports BNT's original random-split
result, Com-BrainTF, and Rahman~\etal; their splits, site subsets,
and metrics prevent direct ranking against the matched results.
Our evaluation includes small sites such as SBL ($n=30$) and
CALTECH ($n=38$).
In Figure~\ref{fig:main_results}C, 12 of 20 sites exceed 0.75 AUROC;
the highest values are UCLA\_2 (0.94), UM\_2 (0.93), and USM (0.88).
NYU, the largest site ($n=184$), achieves 0.782.

\begin{table}[t]
\centering
\caption{ABIDE-I classification across all 20 sites under
\loso{} (mean-site AUROC). $\pm$ denotes five-seed standard
deviation where available.}
\label{tab:classification}
\resizebox{\columnwidth}{!}{%
\begin{tabular}{lccc}
\toprule
Method & Eval & \auroc{} & Acc (\%) \\
\midrule
LR + FC
  & \loso{} & $0.747\pm0.094$ & -- \\
SVM + FC
  & \loso{} & $0.734\pm0.086$ & -- \\
BNT~\cite{kan2022brain}
  & \loso{} & 0.701 & 66.6 \\
\midrule
\ours{} DDPM anomaly
  & \loso{} & $0.718\pm0.013$ & -- \\
\textbf{\ours{} MLP classifier}
  & \loso{} & $\mathbf{0.798}\pm0.003$
  & $\mathbf{71.0}\pm0.4$ \\
\bottomrule
\end{tabular}%
}
\end{table}

\begin{table}[t]
\centering
\caption{Contextual results with different splits, site subsets,
or metrics; these cannot be ranked against Table~\ref{tab:classification}.}
\label{tab:context}
\resizebox{\columnwidth}{!}{%
\begin{tabular}{llcc}
\toprule
Method & Evaluation & AUROC & Acc (\%) \\
\midrule
BNT (original report)~\cite{kan2022brain}
  & Random split & 0.802 & -- \\
Com-BrainTF~\cite{bannadabhavi2023community}
  & Random 70/10/20 & 0.727 & 66.4 \\
Rahman~\etal~\cite{rahman2026multimodal}
  & LOSO, 17 sites & -- & 82.0 \\
\bottomrule
\end{tabular}%
}
\end{table}

\subsection{Matched Healthy-Reference Scorers}
\label{sec:matched}

With the supervised encoder fixed, DDPM achieves $0.718\pm0.013$
mean-site AUROC versus 0.68--0.71 for six healthy-reference alternatives;
normalizing flow is closest at 0.71 (Table~\ref{tab:matched}).
Without paired comparisons and alternative-scorer uncertainty,
these numerical differences do not establish statistical superiority.

\begin{table}[t]
\centering
\caption{Healthy-only scorers on the same supervised
128-dimensional representation (mean-site \loso{} AUROC).
Alternative means are rounded to two decimals; only DDPM includes
a reported five-seed standard deviation.}
\label{tab:matched}
\begin{tabular}{lc}
\toprule
Scorer & AUROC \\
\midrule
Distance to healthy centroid & 0.69 \\
Mahalanobis (single Gaussian) & 0.70 \\
$k$NN distance ($k=10$) & 0.70 \\
One-class SVM & 0.68 \\
VAE reconstruction error & 0.70 \\
Normalizing-flow likelihood score & 0.71 \\
Latent DDPM (ours) & $0.718\pm0.013$ \\
\bottomrule
\end{tabular}
\end{table}

\subsection{Corruption-Step Sensitivity}
\label{sec:sensitivity}

AUROC varies across tested corruption levels, with the highest observed
value at $t^{\ast}=100$ (Table~\ref{tab:sensitivity}).
These descriptive comparisons do not establish a statistically
distinct optimum.

\begin{table}[t]
\centering
\caption{Corruption-step sensitivity (mean-site \loso{}
AUROC); uncertainty in between-level differences is not reported.}
\label{tab:sensitivity}
\begin{tabular}{lccccc}
\toprule
$t^{\ast}$ & 25 & 50 & 100 & 150 & 200 \\
\midrule
AUROC & 0.66 & 0.70 & 0.718 & 0.71 & 0.69 \\
\bottomrule
\end{tabular}
\end{table}

\subsection{Preprocessing Finding: The FC Transpose}
\label{sec:preprocessing}

The ROI files used here store time-series as $T\times N$
(timepoints $\times$ ROIs). Row-wise correlation without transposition
produces $T\times T$ rather than $N\times N$ connectivity.
Correcting this raises AUROC from 0.571 to 0.798
(Table~\ref{tab:transpose}). This 0.227 difference is pipeline-specific
and does not imply the same error in other implementations.

\begin{table}[t]
\centering
\caption{FC transpose correction in our pipeline
(mean-site \loso{} AUROC).}
\label{tab:transpose}
\begingroup
\setlength{\tabcolsep}{3pt}
\begin{tabular}{lc}
\toprule
FC computation & Mean-site \auroc{} \\
\midrule
Incorrect (rows = timepoints) & 0.571 \\
Correct (rows = ROIs; transposed) & 0.798 \\
\bottomrule
\end{tabular}
\endgroup
\end{table}

\subsection{DDPM Anomaly Scores}
\label{sec:anomaly_results}

ASD subjects have higher DDPM scores than controls
(Table~\ref{tab:anomaly}; Figure~\ref{fig:anomaly_dist}):
one-sided Mann--Whitney $U$, $p<10^{-6}$, rank-biserial effect size 0.21.
Table~\ref{tab:anomaly} reports subject-level SDs, distinct from
across-seed SDs in Table~\ref{tab:classification}.

\begin{table}[t]
\centering
\caption{DDPM scores by diagnosis (mean $\pm$ subject-level SD).
The one-sided Mann--Whitney $U$ test evaluates \asd{} $>$ \hc{};
$r_{\mathrm{rb}}$ is rank-biserial correlation.}
\label{tab:anomaly}
\begingroup
\setlength{\tabcolsep}{5.5pt}
\begin{tabular}{lcccc}
\toprule
Group & $n$ & Score (mean $\pm$ std) & $p$-value
  & $r_{\mathrm{rb}}$ \\
\midrule
\hc{} & 571 & $4.98\pm1.79$ & -- & -- \\
\asd{} & 531 & $6.42\pm3.59$ & $<10^{-6}$ & 0.21 \\
\bottomrule
\end{tabular}
\endgroup
\end{table}

\noindent\textbf{Age stratification.}
ASD mean scores exceed HC within all displayed age groups
(6--12, 12--18, 18--30, 30+; Figure~\ref{fig:main_results}D).
This descriptive stratification does not rule out age confounding.

\subsection{Clinical Severity Correlations}
\label{sec:clinical}

Both DDPM scores and logits show weak positive associations with
ADOS Total and negative associations with full-scale IQ
(Table~\ref{tab:clinical}). DDPM Spearman coefficients are 0.108
and $-0.138$, respectively; after logit adjustment, they are 0.105
and $-0.115$. The partial ADOS association has $p=0.045$, with a
lower CI endpoint rounding to zero; the partial IQ $p$-value is
unreported. Pearson correlations are nonsignificant, but differing
Pearson/Spearman significance does not establish nonlinearity.
These exploratory associations do not establish incremental
prediction or clinical utility; covariate limitations are stated below.

\begin{table*}[t]
\centering
\caption{Within-ASD associations: ADOS Total, $n=375$; IQ,
$n=498$. Coefficients and CI endpoints are rounded to two decimals;
precise DDPM coefficients appear in the text. The partial row controls
for classifier logit only; a dash denotes an unreported $p$-value.}
\label{tab:clinical}
\begin{tabular}{lcccccc}
\toprule
& \multicolumn{3}{c}{ADOS Total}
& \multicolumn{3}{c}{Full-scale IQ} \\
\cmidrule(lr){2-4}\cmidrule(lr){5-7}
Predictor & Correlation & 95\% CI & $p$
  & Correlation & 95\% CI & $p$ \\
\midrule
Classifier logit
  & 0.12 & [0.02, 0.22] & 0.02
  & $-0.14$ & [$-0.23$, $-0.05$] & 0.002 \\
DDPM anomaly score
  & 0.11 & [0.01, 0.21] & 0.038
  & $-0.14$ & [$-0.22$, $-0.05$] & 0.002 \\
DDPM score, partial
  & 0.11 & [0.00, 0.21] & 0.045
  & $-0.12$ & [$-0.20$, $-0.03$] & -- \\
\bottomrule
\end{tabular}
\end{table*}

\subsection{Ablation Study}
\label{sec:ablation}

On the paired subset (Table~\ref{tab:ablation}), structural-only
classification achieves 0.545 AUROC; fusion adds 0.002 over FC alone,
which does not establish added value.
Reconstruction-trained AE+DDPM achieves approximately 0.51 AUROC.
This ablation assesses representation learning; Table~\ref{tab:matched}
separately compares scorers with the supervised encoder fixed.

\begin{table}[t]
\centering
\caption{Mean-site \loso{} ablations. Top: paired fMRI/sMRI
subset ($n=980$); its FC-only AUROC (0.750) is not directly comparable
to the full 1,102-subject result (0.798; Table~\ref{tab:classification}).
Bottom: anomaly scoring with a reconstruction-trained encoder.}
\label{tab:ablation}
\begin{tabular}{lc}
\toprule
Method & \auroc{} \\
\midrule
\multicolumn{2}{l}{\textit{Modality (supervised MLP classifier)}} \\
sMRI only (BrainIAC~\cite{tak2026generalizable}) & 0.545 \\
fMRI FC only & 0.750 \\
fMRI FC + sMRI (\vit{}) & 0.752 \\
\midrule
\multicolumn{2}{l}{\textit{Anomaly scoring (AE representation)}} \\
Unsupervised AE + \ddpm{} & ${\sim}0.51$ \\
\bottomrule
\end{tabular}
\end{table}

\subsection{Independent Exploratory Structural Visualization}
\label{sec:maps}

Group maps from the IXI-only pretrained 3D \vit{}
(Figure~\ref{fig:brain_maps}) qualitatively suggest higher ASD
reconstruction errors in frontal and temporoparietal regions;
selected individual maps show heterogeneous patterns
(Figure~\ref{fig:individual_maps}).
These separate structural maps neither contribute to nor spatially
explain the fMRI DDPM score and do not establish regional biomarkers.

\subsection{Discussion}
\label{sec:discussion}

The MLP's higher diagnostic AUROC is consistent with its classification
objective. DDPM measures reconstruction-based deviation from a healthy
reference within the ASD-supervised representation; this is not a
label-free pipeline. Classifier logits are also continuous, and weak
logit-adjusted associations do not demonstrate a benefit from combining
the scores. Across-seed SDs summarize complete runs and do not isolate
stochastic reconstruction variability with fixed model parameters.

\subsection{External Dataset Validation}
\label{sec:external}

Before external testing, hyperparameters were fixed using ABIDE-I
\loso{} experiments; the final encoder and DDPM were trained on all
ABIDE-I subjects and healthy controls, respectively, then frozen.
Normalization parameters were fitted on ABIDE-I and applied unchanged,
with no external-cohort retraining or tuning.
The ABIDE-II~\cite{di2017enhancing} evaluation uses 847 subjects from
17 sites with the same CC200 parcellation; ADHD-200~\cite{adhd2012adhd}
evaluation uses 362 ADHD subjects and 585 controls.
Frozen DDPM pooled AUROCs are 0.69 and 0.61, respectively, versus
0.74 and 0.63 for the classifier (Table~\ref{tab:external}).
Transfer remains preliminary, particularly to ADHD given the
ASD-supervised encoder.

\begin{table}[t]
\centering
\caption{ABIDE-I: mean-site \loso{} AUROC $\pm$ five-seed SD.
External cohorts: pooled AUROC using frozen ABIDE-I-trained models.
The aggregations differ. $^\dagger$Exploratory cross-disorder transfer
with an ASD-supervised encoder.}
\label{tab:external}
\resizebox{\columnwidth}{!}{%
\begin{tabular}{llcc}
\toprule
Dataset & Disorder & MLP & DDPM \\
\midrule
ABIDE-I (internal)
  & ASD & $0.798\pm0.003$ & $0.718\pm0.013$ \\
ABIDE-II (external)
  & ASD & 0.74 & 0.69 \\
ADHD-200 (external)$^\dagger$
  & ADHD & 0.63 & 0.61 \\
\bottomrule
\end{tabular}%
}
\end{table}

\noindent\textbf{Limitations.}
Scorer comparisons are conditional on supervised representations.
Paired tests and alternative-scorer uncertainty are not reported;
inference-time stability and combined-score predictive benefit
remain unestablished. Both diagnostic-group and clinical analyses
lack adjustment for age, sex, head motion, and site; clinical analyses
also use subsets with available measurements. Weak associations do
not establish severity grading, clinical utility, or causality, and
scores are not calibrated disease probabilities.
Healthy-reference representativeness and residual site effects remain
concerns (MLP site AUROC approximately 0.68--0.94); CC200 limits spatial
detail. Small structural-fusion differences do not establish added
value, and external transfer does not establish disorder specificity.
Future work includes paired scorer comparisons, stochastic stability,
combined-score prediction, broader healthy references, alternative
parcellations, covariate adjustment, calibration, and prospective
clinical utility.

\subsection{Future Methodological Extensions}
\label{sec:future_extensions}

Beyond the statistical and clinical validation priorities identified
above, several methodological directions could extend the present
framework. These directions would require additional implementation,
data, and evaluation.

\noindent\textbf{Adaptive computation and optimization.}
The encoder and diffusion noise predictor both use MLP components,
making adaptive computation a potential architectural extension.
MID-L~\cite{shaeri2025mid} investigates input-dependent neuron selection
through learned gating and matrix-interpolated transformation paths.
Future work could examine whether such selection reduces the
computational cost of connectivity encoding or noise prediction.
Tri-Accel~\cite{sheibanian2025tri} provides a complementary optimization
direction through curvature-informed precision adaptation and
memory-aware batch sizing. Evaluating these approaches within \ours{}
would require measuring computational cost alongside cross-site
AUROC and stochastic anomaly-score stability.

\noindent\textbf{Modular data curation and evaluation.}
MNIST-Gen~\cite{shaeri2025mnist} organizes task-specific image-dataset
construction through hierarchical semantic categorization and modular
processing stages. A corresponding design direction for neuroimaging
is to organize preprocessing, quality review, cohort metadata, and
evaluation subsets as explicit, traceable components. Such a workflow
could support systematic audits across acquisition sites, demographic
groups, and preprocessing configurations, extending the reproducibility
checks illustrated by our FC-orientation analysis.

\noindent\textbf{Joint prediction and contextual information.}
ClimateEmote~\cite{shaericlimateemote} studies sentiment, emotion,
misinformation, and stance prediction within a multi-task framework.
This provides a cross-domain motivation for investigating shared
representations with task-specific objectives. A future extension of
\ours{} could jointly study diagnostic classification, healthy-reference
deviation, and clinical-measure prediction, while assessing each
objective separately. The survey of sentiment and social signals during
extreme weather events~\cite{shaeri2025sentiment} additionally examines
methods for extracting contextual information from heterogeneous
language data. In a future cohort containing suitable participant
reports, analogous text-processing components could be investigated
alongside imaging representations to study relationships between
reported experiences and neuroimaging-derived deviations.

\noindent\textbf{Question-guided evidence review.}
Beigi~\etal~\cite{beigi2025can} investigate relevant question generation
to support evidence retrieval in multimodal fact-checking. This
suggests a potential interface direction for reviewing anomaly scores:
structured questions could organize checks of acquisition quality,
head motion, reference-population coverage, and agreement between
functional and structural measurements. Such an extension would
require evaluating whether retrieved evidence supports the generated
questions and whether the resulting workflow improves expert review.

\noindent\textbf{Domain-informed multimodal integration.}
Environmental modeling offers examples of combining heterogeneous
measurements with domain-specific structure.
WebMRT~\cite{alkhaled2024webmrt} uses machine learning for online
mean-radiant-temperature prediction, while the multimodal
physics-informed approach of Shaeri~\etal~\cite{shaeri2025multimodal, shaeri2025saud}
combines environmental inputs with radiation-based physical constraints.
OpenMRT~\cite{middel2026openmrt} and
VisionMRT~\cite{middel2026visionmrt} investigate thermal-exposure
modeling from imagery, with VisionMRT incorporating 3D reconstruction.
OpenIoTwx~\cite{Shaeri2026} contributes a modular platform for
environmental measurements. Together, these works motivate a broader
design pattern that separates measurement collection, modality-specific
processing, and domain-informed integration. For future neuroimaging
extensions of \ours{}, this pattern could guide the combination of
functional connectivity, structural MRI, physiological signals, and
acquisition metadata, using constraints justified specifically for
those measurements. The contribution of each additional modality or
constraint would require matched ablations and external validation.
\section{Conclusion}
\label{sec:conclusion}

We presented \ours{}, a healthy-reference latent diffusion framework
for anomaly scoring in ASD using diagnostically supervised
functional-connectivity representations.
Healthy-only DDPM fitting provides a continuous deviation score,
while ASD/HC labels structure the encoder's latent space.
Across 20 ABIDE-I sites and five seeds, the classifier achieves
a mean-site AUROC of $0.798\pm0.003$ and the DDPM score
$0.718\pm0.013$. Matched healthy-only scorers reach 0.68--0.71; the DDPM's modest
numerical advantage is conditional on the supervised representation
and does not establish statistical superiority.
Among tested corruption levels, the highest observed AUROC
occurs at $t^{\ast}=100$. Clinical associations are weak and exploratory after
controlling for classifier logit, without adjustment for age, sex,
site, or head motion. These associations do not establish
incremental predictive or clinical value.
Frozen DDPM evaluation yields pooled AUROCs of 0.69 on ABIDE-II
and 0.61 on ADHD-200, providing preliminary evidence of cross-cohort
and cross-disorder transfer; clinical utility and disorder specificity
remain unresolved. Independent structural maps illustrate exploratory reconstruction
differences, and the FC-orientation check highlights a
reproducibility issue within this pipeline.
These results motivate broader healthy references, alternative
parcellations, covariate-adjusted analyses, cohort-specific validation,
and evaluation of calibration and clinical utility.
{\small
\bibliographystyle{ieeenat_fullname}
\bibliography{main}
}
\end{document}